%% file: main.tex
\documentclass[runningheads]{llncs}

\usepackage{eccv}

\usepackage{eccvabbrv}

\usepackage{graphicx}
\usepackage{booktabs}

\usepackage[accsupp]{axessibility}  % Improves PDF readability for those with disabilities.
\usepackage{algorithm,algorithmic}% http://ctan.org/pkg/algorithms

\usepackage{algorithm}
\usepackage{listings}
\usepackage{amsmath}
\usepackage{colortbl} % This package is needed for coloring table rows
\usepackage{pifont}

\newcommand{\xmark}{\ding{55}}
\usepackage{mathtools}
\DeclarePairedDelimiter\floor{\lfloor}{\rfloor}

\usepackage{titletoc}

\usepackage{hyperref}

\usepackage{orcidlink}

\begin{document}

\input{1.main_contents}

% ---- Bibliography ----
%
% BibTeX users should specify bibliography style 'splncs04'.
% References will then be sorted and formatted in the correct style.
%
\bibliographystyle{splncs04}
\bibliography{main}

\input{2.supp}

\end{document}

%% file: 1.main_contents.tex
% ---------------------------------------------------------------
\title{Memory-Efficient Fine-Tuning \\ for Quantized Diffusion Model} 

\titlerunning{TuneQDM}

\author{Hyogon Ryu \and
Seohyun Lim \and
Hyunjung Shim}

\authorrunning{Ryu et al.}

\institute{Korea Advanced Institute of Science and Technology (KAIST)\\
\email{\{hyogon.ryu, seohyunlim, kateshim\}@kaist.ac.kr}}

\maketitle
\begin{abstract}
The emergence of billion-parameter diffusion models such as Stable Diffusion XL, Imagen, and DALL-E 3 has significantly propelled the domain of generative AI. However, their large-scale architecture presents challenges in fine-tuning and deployment due to high resource demands and slow inference speed. This paper explores the relatively unexplored yet promising realm of fine-tuning quantized diffusion models. Our analysis revealed that the baseline neglects the distinct patterns in model weights and the different roles throughout time steps when finetuning the diffusion model. 
To address these limitations, we introduce a novel memory-efficient fine-tuning method specifically designed for quantized diffusion models, dubbed TuneQDM. Our approach introduces quantization scales as separable functions to consider inter-channel weight patterns. Then, it optimizes these scales in a timestep-specific manner for effective reflection of the role of each time step. TuneQDM achieves performance on par with its full-precision counterpart while simultaneously offering significant memory efficiency. Experimental results demonstrate that our method consistently outperforms the baseline in both single-/multi-subject generations, exhibiting high subject fidelity and prompt fidelity comparable to the full precision model.
  \keywords{Quantization \and Diffusion Model \and Transfer Learning}
\end{abstract}

\section{Introduction}
\label{sec:intro}

Diffusion models have been a de facto standard in generative models, especially in image synthesis~\cite{rombach2022high, saharia2022photorealistic, dhariwal2021diffusion, ho2020denoising, nichol2021improved}. They are widely used in various applications, such as image super-resolution~\cite{saharia2022image, li2022srdiff}, inpainting~\cite{yeh2017semantic, lugmayr2022repaint}, and text-to-image generation~\cite{gal2022image, ding2022cogview2, rombach2022high, ruiz2023dreambooth, balaji2022ediffi}. However, their slow generation process and high memory and computational requirements pose significant challenges for practical use. 

With the emergence of billion-parameter diffusion models such as Stable Diffusion XL~\cite{podell2023sdxl}, Imagen~\cite{saharia2022photorealistic}, and DALL-E 3~\cite{betker2023improving}, the issues of slow inference and computational load are becoming more pronounced. Recent studies have focused on model quantization to address these concerns. Quantization~\cite{li2016ternary, han2015deep, jacob2018quantization, jung2019learning, shang2023post, li2023q} is a key model compression technique that uses lower-bit representations (e.g., 4-bit, 8-bit) for model parameters, thus drastically improving computational and memory efficiency. Notably, PTQ4DM~\cite{shang2023post} has achieved 8-bit quantization in diffusion models by constructing a timestep-aware calibration dataset and Q-Diffusion~\cite{li2023q} has accomplished both 8-bit and 4-bit quantization by separating the shortcut layer through activation analysis.

Given the growing role of diffusion models as vision foundation models, the direct fine-tuning of quantized diffusion models for specific applications is an unexplored yet highly impactful research direction. This approach is inspired by recent developments in the large language model (LLM), where techniques like Alpha Tuning, PreQuant, and PEQA~\cite{kwon2022alphatuning, gong2023prequant, kim2024memory} have been investigated for fine-tuning quantized LLMs. 

\begin{figure}[tb]
  \centering
  \includegraphics[width=\linewidth]{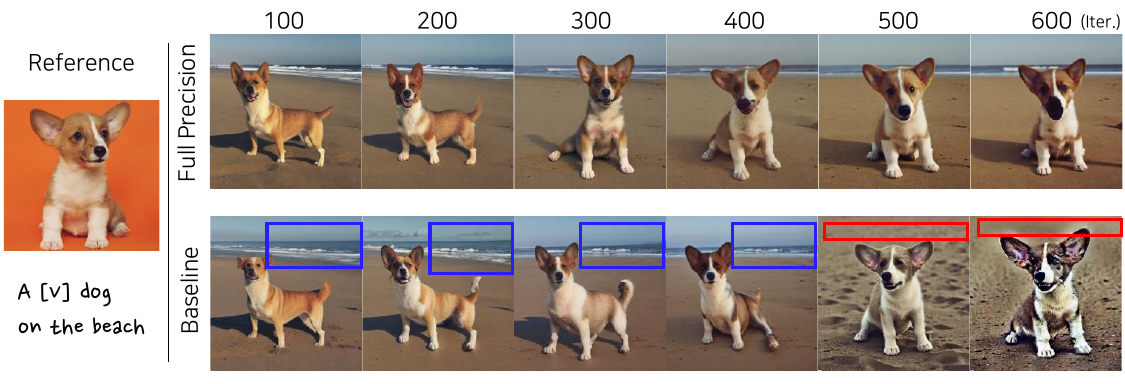}
  \caption{\textbf{Comparison between fine-tuning a full precision model and fine-tuning a quantized diffusion model with the baseline}. Unlike the \texttt{fp} model, the \texttt{baseline} cannot achieve both prompt fidelity and subject fidelity simultaneously. Up to 400 iterations, it retains high image quality but fails to accurately reflect reference features. After 500 iterations, the ocean disappears, and further training leads to noticeable artifacts. Blue boxes indicate where the ocean is present, while red boxes highlight areas where the ocean should be but is missing. A unique token, \texttt{[V]}, is used as an identifier describing images provided by users.
  }
  \label{fig:baseline-limitation-1}
\end{figure}

Building on the success of the LLM community, we developed a baseline for fine-tuning quantized diffusion models. Leveraging publicly available quantized checkpoints from Q-Diffusion, we trained the model using the PEQA methodology, commonly used for fine-tuning quantized LLMs. For fine-tuning diffusion models, we utilized the common diffusion personalization technique, DreamBooth. Combining these three cutting-edge methods, we established a baseline and observed its performance trend. As seen in Fig.~\ref{fig:baseline-limitation-1}, the \texttt{baseline} model produced unsatisfactory results in terms of either fidelity or quality. The generated results either fail to reflect the concept of text prompt (e.g., ocean) or personalized categories denoted by an identifier token \texttt{[V]}. Moreover, image quality degraded with more training iterations.

\begin{figure}[tb]
  \centering
  \includegraphics[width=\linewidth]{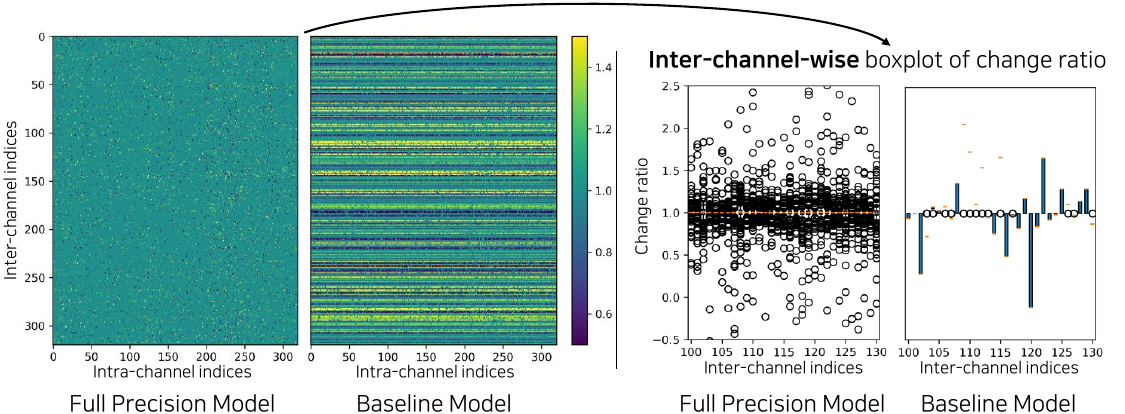}
  \caption{\textbf{Weight change ratio after the fine-tuning}. The left side describes the weight change ratio of \texttt{fp} model and \texttt{baseline} in 2D image plots, and the right side describes it in an inter-channel-wise boxplot. There is a clear difference between the \texttt{baseline} and the \texttt{fp} model.
  }
  \label{fig:baseline-limitation-2}
\end{figure}

To understand the performance limitations, we first fine-tune the model and compare the weight update patterns before and after quantization. Interestingly, the distinct inter-channel patterns appear in the change ratio map between the full-precision (\texttt{fp}) weight matrix and its fine-tuned version. Fig. \ref{fig:baseline-limitation-2} illustrates the ratio map, highlighting the changes made by fine-tuning. Ideally, the ratio map from the full-precision model should be the target. However, after quantization, these maps are clearly different from those of the full-precision model. Inter-channel patterns disappear in the \texttt{baseline} because the scale parameters, the only parameter trainable in the \texttt{baseline}, vary only within the channel. As observed in Fig. \ref{fig:baseline-limitation-2}, the \texttt{fp} model shows distinct inter-channel patterns in weight change ratio while the \texttt{baseline} fails to capture these weight updates effectively.
 
Secondly, we focus on timestep-aware training for fine-tuning quantized diffusion models. Previous studies \cite{choi2022p2weight, balaji2022ediffi, lee2023meme} discuss the significant role of timesteps in diffusion training. It is known that during the denoising step of diffusion models, intervals affecting content features differ from those affecting coarse features or noise-cleaning. Therefore, we designed a method to separate the timesteps and assign different roles to each timestep during fine-tuning. 
 
To address these challenges, we propose a novel method for fine-tuning quantized diffusion models called TuneQDM. Our method tackles the aforementioned issues by (1) decomposing the quantization scales as separable functions to consider inter-channel weight patterns and (2) fine-tuning these scales timestep-wisely to reflect the role of the timestep. TuneQDM improves personalization performance while preserving the memory and computational efficiency of quantized diffusion models.

We evaluated our method on personalization (\ie single-/multi-subject generation) and unconditional generation. The results demonstrate that our fine-tuned model can generate images at a level comparable to full precision fine-tuning on both personalization and unconditional generation. Compared to the baseline, TuneQDM addresses issues such as degradation of text prompt and subject fidelity. Particularly, even when fine-tuning the 4-bit ($ 8\times$ compressed) model, we achieved performance levels similar to those of the full precision model. 

In summary, our contributions are as follows: (1) We established a strong baseline by combining existing state-of-the-art methods, identifying limitations, and observing performance trends. (2) We introduced a novel memory-efficient fine-tuning method for quantized diffusion models named TuneQDM. By introducing a multi-channel-wise scale update, we addressed the issue of inter-channel patterns during weight updates. Additionally, by fine-tuning independent scale parameters for each timestep interval, we enabled the quantized diffusion model to effectively reflect the role of each timestep interval. (3) TuneQDM achieves both parameter efficiency and a substantial reduction in memory footprint during fine-tuning. Experimental results demonstrate that our method consistently outperforms the baseline in the single-/multi-subject generation, achieving high subject fidelity and prompt fidelity comparable to the full precision model.

\section{Related work}
\subsection{Quantizing diffusion model}

Quantization reduces model complexity and enhances speed by representing model weights with fewer bits. The two main approaches in quantization are Quantization Aware Training (QAT, integrated during training)\cite{jung2019learning, esser2019learned, jacob2018quantization}, and Post-Training Quantization (PTQ, applied after model training)\cite{banner2019post, nagel2019data, xiao2023smoothquant, hubara2021accurate, yao2022zeroquant, li2023q, shang2023post, he2024ptqd}.

Quantizing diffusion models tends to align well with PTQ\cite{li2023q, shang2023post, he2024ptqd} because diffusion models often serve as foundation models pre-trained on extensive datasets, and retraining the entire model involves high computational overheads. PTQ4DM \cite{shang2023post} constructed a calibration dataset by considering the multi-step process of the diffusion model, and Q-Diffusion \cite{li2023q} proposed a quantization method that takes into account the shortcut connections in the UNet. PTQD \cite{he2024ptqd} decomposed the quantization noise and corrected it. However, research on fine-tuning quantized diffusion models for downstream tasks has not yet been conducted, and we are the first to fine-tune the quantized diffusion model for the downstream task.

\subsection{Personalizing diffusion model}

Text-to-image (T2I)\cite{
saharia2022photorealistic, ding2022cogview2, ramesh2021zero, betker2023improving, podell2023sdxl} generation has garnered significant attention for its ability to produce diverse and realistic images in response to textual prompts. While large models trained on extensive text and image-paired datasets excel in general tasks, they often face difficulties in generating highly personalized or novel images aligned with specific user concepts. Personalization emerges as a prominent downstream task for general diffusion models, aiming to tailor the models to individual preferences or user-defined concepts for image generation.

The users provide several image examples representing the personal concept, while additional scene components, such as backgrounds or attributes, are defined through textual prompts. Textual Inversion\cite{gal2022image} proposed an optimization approach for word embeddings that effectively represents a given image. Meanwhile, DreamBooth \cite{ruiz2023dreambooth} employs the strategy of fine-tuning a pre-trained model to generate images with a novel perspective of the input target. Custom Diffusion\cite{kumari2023customdiffusion} introduces the fine-tuning of only the key and value components of the cross-attention layer, enabling multi-subject image generation.

\subsection{Parameter-efficient fine-tuning}
Fine-tuning enables pretrained models to be adapted to specific tasks. However, full model fine-tuning requires significant computational resources. As an alternative to full fine-tuning, parameter-efficient fine-tuning methods have been proposed, where most of the model's parameters are frozen, and only a subset is updated. Adapter modules\cite{houlsby2019parameter, lin2020exploring, rebuffi2017learning, ruckle2020adapterdrop}  suggest inserting task-specific parameters within pretrained model layers. LoRA\cite{hu2021lora, Ryu_Low-rank_adaptation_for} represents the gap between fully fine-tuned weights and pretrained weights as low-rank matrices, allowing the addition of trainable weights for task adaptation while preserving the pre-trained weights.

However, parameter-efficient fine-tuning methods still struggle to handle a vast number of parameters and are less suitable for scenarios requiring smaller model sizes, such as low-power mobile devices\cite{wu2020easyquant, przewlocka2022power}. Recently, methods for fine-tuning quantized models have been proposed. In the field of LLMs, QLoRA\cite{dettmers2024qlora} proposed a low-rank adaptor applicable to quantized LLMs, while PEQA\cite{kim2024memory} suggested updating scale parameters while freezing quantized weights. Prequant\cite{gong2023prequant} and OWQ\cite{lee2023owq} performed task-agnostic quantization followed by tuning a small subset of weights. Our method applies this concept to fine-tuning diffusion models and demonstrates its effectiveness in personalization and unconditional generation.

%-------------------------------------------------------------------------

\section{Motivation }\label{sec:motivation}

The recent success of large-scale foundation models has led to their widespread adoption in numerous downstream tasks. In the field of computer vision, the diffusion model has emerged as a representative foundation model and is popularly used in various fields such as personalized generation, 3D generation combined with NeRF\cite{mildenhall2021nerf}, and improving discriminative model training, leveraging models like Stable Diffusion\cite{rombach2022high} or Imagen\cite{saharia2022photorealistic}. As foundation models expand exponentially in scale over time, it has led to a growing interest in fine-tuning reduced (a.k.a., quantized) foundation models for specific downstream tasks. This concept has been actively explored in the natural language processing field using LLMs. In this study, we are the first to apply the same philosophy to the vision foundation model, the diffusion model. We specifically introduce a new problem of directly applying quantized diffusion models to a key downstream task--personalization.

\begin{figure}[tb]
  \centering
  \includegraphics[width=\linewidth]{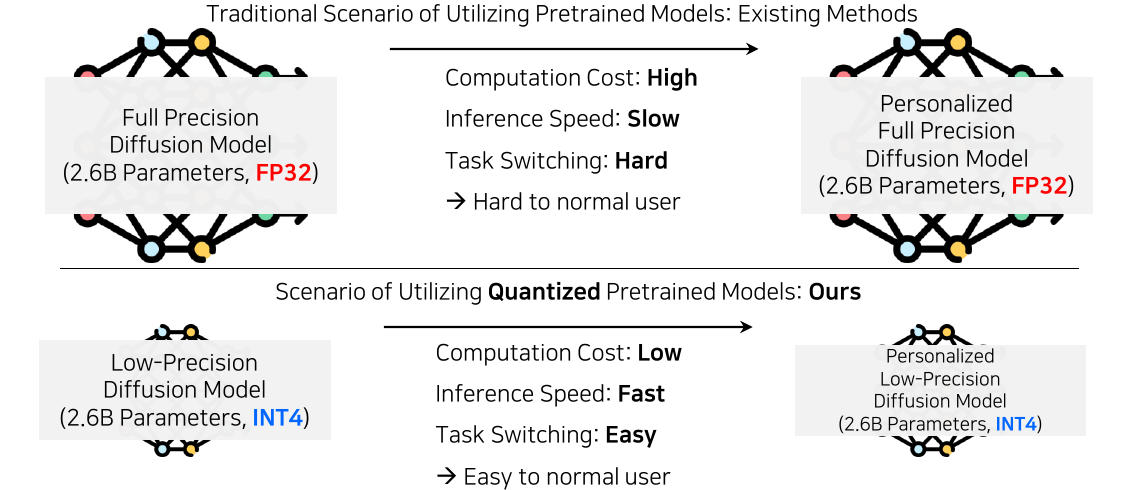}
  \caption{\textbf{Scenario of utilizing quantized pretrained models.} 
Above: Full model is loaded and fine-tuned. Below: Quantized model is loaded and used. As the model sizes increase, fine-tuning requires significant computational cost. Therefore, directly fine-tuning the quantized model offers various efficiency advantages for users.
  }
  \label{fig:overall-concept}
\end{figure}

\noindent\textbf{Advantages of fine-tuning quantized diffusion model.} Diffusion models have increasingly utilized larger UNets to improve image quality. For example, the Stable Diffusion model has expanded to 2.6 billion parameters in its SDXL variant. Similarly, Imagen's model has grown to 3 billion, while DALL-E 2 has escalated to 5.5 billion parameters. This upward trend in the sizes of foundation models raises concerns about the efficiency of deploying and utilizing pretrained diffusion models.

To align with this trend, we introduce the concept of fine-tuning quantized diffusion models. As described in Fig \ref{fig:overall-concept}, directly fine-tuning quantized diffusion models for downstream tasks offers several advantages. First, quantized checkpoints require significantly less memory storage, DRAM, and fewer trainable parameters than their full-precision weights. Besides, we store only the scale parameters per dataset ($\sim$ 3MB) and reuse the quantized checkpoint across different datasets. This approach eliminates the need for separate quantization processes (\ie, PTQ) for each dataset and task, which is particularly time-consuming for diffusion models. Finally, the computational costs for deployment are substantially reduced compared to their full-precision counterparts.

%-------------------------------------------------------------------------
\section{Methodology}
In this section, we propose a memory-efficient fine-tuning method using quantized diffusion models. We first provide a brief overview of uniform quantization and introduce the baseline using the PEQA technique. Then, we analyze the challenges of the baseline method and propose our solutions. Finally, we introduce TuneQDM.

\subsection{Post-training quantization} \label{sec:preliminary}

Post-training quantization is the prevalent method to quantize the model weights with low precision. In this paper, we utilize the hardware-friendly quantization method, uniform quantization. For a given full-precision model's weight matrix $W_f$, the quantization and de-quantization operation can be expressed as:

\begin{align}
    W_q &= \texttt{clamp} \big( \texttt{round} \big( \frac{W_f}{s}\big) + z, 0, 2^b - 1\big), \\
    \hat{W}_f &= s \cdot ( W_q - z ),
\end{align}
\noindent where $W_q$ and $\hat{W}_f$ represent the quantized weight indices and de-quantized weight matrix, respectively. Here, $s$, $z$, and $b$ are per-channel scaling factors, zero points, and number of bits, respectively. The function $\texttt{round}(\cdot)$ and $\texttt{clamp}(\cdot, a, b)$ are used for rounding and clamping within the range $[a,b]$.

\subsection{Baseline} \label{sec:baseline}
We have constructed a baseline by applying PEQA, a method for fine-tuning quantized LLMs, to quantized diffusion models. PEQA involves freezing the weight integers in the quantized pretrained model and updating only the quantization scale for fine-tuning. Through PEQA, the fine-tuned weight $W_{tuned}$ can be expressed as follows:

\begin{align}
    W_{tuned} &= (s + \Delta s) \cdot (W_q - z) \\
    &= (s + \Delta s) \cdot \big( \texttt{clamp} \big( \texttt{round} \big( \frac{W_f}{s} \big) + z, 0, 2^b - 1\big) - z \big).
\end{align}

Here, only $s$ is trainable, while the other parameters remain frozen. The weight indices, scale, and zero-point parameters are initialized using the quantized diffusion model checkpoint (\ie, Q-Diffusion).

\noindent\textbf{Limitation of baseline}. During our experiments with the baseline, we identified two key issues in the fine-tuning process of the quantized model: (\textbf{P1}) \textbf{Weight update across channels:} As depicted in Fig. \ref{fig:baseline-limitation-2}, weight updates during the fine-tuning should occur independently of the channels. However, in the baseline approach, only the intra-channel-wise scale parameters are updated, restricting changes in the inter-channel components of the weight matrix. This limitation is observed consistently in both linear and conv2d layers. (\textbf{P2}) \textbf{Limited capacity to learn denoising timestep variability:} The role of the UNet architecture in the diffusion model varies depending on the denoising timestep. However, we found that the baseline's capacity to learn these variations is limited. Even with extended training iterations, the output images that are not exactly the same with target images. There are often some quality degradation or difference in details.

% ---------
\subsection{ Multi-channel-wise scale update}
To address the aforementioned issue (\textbf{P1}), we propose the multi-channel-wise scale update method. Given the quantized weight $W_q \in \mathbb{R}^{n \times m}$ (for conv2d layers, $W_q \in \mathbb{R}^{n \times m \times k \times k}$), and the intra per-channel quantization scale $s_{\text{out}} \in \mathbb{R}^m$, we define the inter per-channel quantization scale $s_{\text{in}} \in \mathbb{R}^n$. Subsequently, while freezing the quantized integer values $W_q$, we only fine-tune $s_{\text{in}}$ and $s_{\text{out}}$. The fine-tuned weights $W_{\text{tuned}}$ for downstream task are expressed as:

\begin{equation}
 W_{\text{tuned}} = (s_{\text{out}} + \Delta s_{\text{out}}) \cdot (W_q^* - z^*) \cdot (s_{\text{in}} + \Delta s_{\text{in}}). 
\end{equation}
 
\noindent Here, $\Delta s_{\text{in}} \in \mathbb{R}^n$ and $\Delta s_{\text{out}} \in \mathbb{R}^m$ represent the gradient updates generated during fine-tuning for the downstream task. * indicates the frozen parameters.

\begin{figure}[t]
  \centering
  \includegraphics[width=\linewidth]{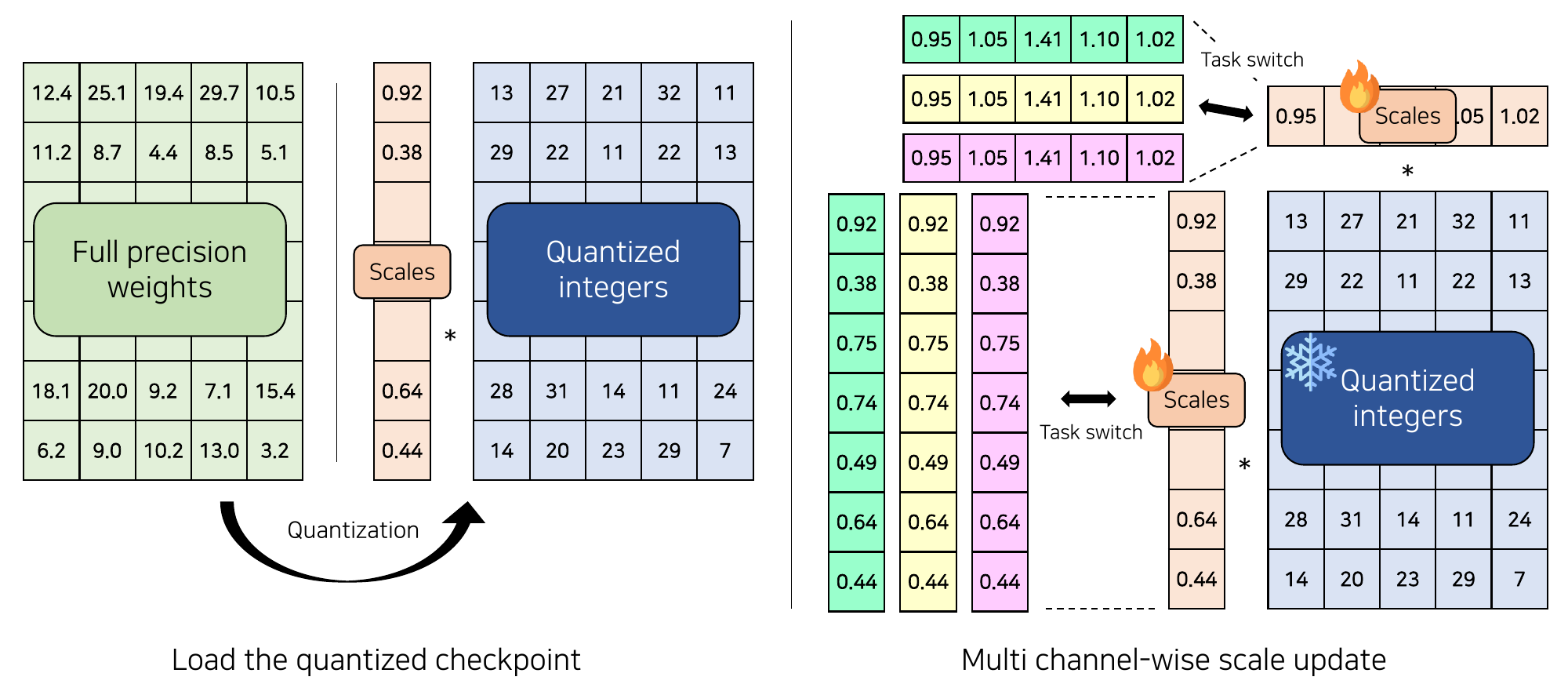}
  \caption{\textbf{Multi-channel-wise-scale}. Left: Our method requires quantization to be performed only once on a pretrained model, enabling subsequent fine-tuning across various tasks without large computation costs. Right: When switching tasks, the scale pairs should be switched together. This simplifies task switching and allows the quantized model to be easily adapted to different tasks.
  }
  \label{fig:multi-channel-wise-scale}
\end{figure}

The multi-channel-wise scale update is a memory-efficient fine-tuning method directly applicable to quantized diffusion models. It requires only $(m+n)$ trainable parameters, as opposed to the full weight matrix $W_f$, making it parameter-efficient. Moreover, as $W_q$ and $z$ remain fixed, adjusting only $(s_{\text{out}} + \Delta s_{\text{out}})$ and $(s_{\text{in}} + \Delta s_{\text{in}})$ values allows for easy application to other downstream tasks, facilitating task switching. While the baseline represents a specific case with all elements of $s_{\text{in}}$ being 1 and frozen, our method can be considered as a generalized approach. The overall process of the multi-channel-wise scale update is illustrated in Fig \ref{fig:multi-channel-wise-scale}.

\subsection{Timestep-aware scale update strategy}

To address the issue (\textbf{P2}), we propose a timestep-aware scale update strategy. P2weighting\cite{choi2022p2weight} elucidates the varied contributions of training timesteps to image generation. Similarly, e-Diffi\cite{balaji2022ediffi} and MEME\cite{lee2023meme} enhance text-to-image generation performance by employing multi-expert models that adapt based on the training timesteps. 

Inspired by the above-mentioned approaches, we independently fine-tune quantization scales for each timestep to update multi-experts. Assuming we propose $n$ experts, we uniformly divide the timestep interval into $n$ segments. For each segment, we clone and update quantization scales separately. After training, we obtain $n$ optimized quantization scales $S_i$ for each timestep interval $I_i$:
\begin{equation}
 S_n = \{ s_1, s_2, …, s_n \}, \quad \mathcal{I}_n = \{I_i | ( \frac{i \times t}{n}, \frac{(i+1) \times T}{n})  \} \quad \text{ for } i=1\ldots n,
\end{equation}
where $T$ is the total denoising steps. Our approach is much more memory-efficient compared to conventional methods that train full models to create multi-experts because we only update scale parameters. During inference, the quantized integer values remain fixed, and only the scale parameters switch according to each timestep.

\begin{algorithm}
\caption{TuneQDM pipeline}
\begin{algorithmic}[1]
    \REQUIRE  Quantized model weight set $\{W_q^{(l)}\}_{l=1}^L$, Number of layer $L$
    \REQUIRE Quantization parameter set $\{(\textbf{s}, \textbf{z})_l\}_{l=1}^L$
    \REQUIRE Number of expert $N$ and number of denoising step $T$
    \REQUIRE Training dataset $\{(x,c,t)\}_{m=1}^M$, x, c, t are the noisy image, condition timestep, repectively.

    \FOR {$l = 1 : L$}
        \FOR{$n = 1 : N$}
            \STATE Initialize intra-channel scale $\textbf{s}^{intra} \sim \mathcal{N}(1, 0.01)$
        \ENDFOR
    \ENDFOR

    \STATE Freeze $W_q$ and set only $\textbf{s}$ and $\textbf{s}^{intra}$ are trainable
    
    \FOR {\textit{training epoch}}
        \FOR{ $(x,c,t)$ \textit{in train dataset} }
            \STATE \textit{Expert index} $i \gets \floor{ \frac{t*N}{T} }$
            \STATE Update $i$\textit{-th expert} $\textbf{s}$ and $\textbf{s}^{intra}$ for each layer $l$.
        \ENDFOR
    \ENDFOR
    
    \RETURN Fine-tuned inter/intra-channel scale parameters $\{(\textbf{s}, \textbf{s}^{intra})_l\}_{l=1}^{L}$
\end{algorithmic}
\label{alg:TuneQDMpipe}
\end{algorithm}

\subsection{Memory-efficient fine-tuning for quantized diffusion model}
We propose TuneQDM, a novel fine-tuning method for quantized diffusion models. The overall pipeline is outlined in Algorithm \ref{alg:TuneQDMpipe}. This method supports the previously introduced multi-channel-wise scale update and timestep-aware scale update strategies.

Firstly, we initialize the weights and quantization parameters using the quantized diffusion checkpoint and initialize the multi-channel-wise scales. Then, we fine-tune the quantized model with the downstream task's loss function. If the timestep-aware scale update strategy is employed, additional training is conducted according to the timestep interval. 

This systematic approach integrates key strategies such as multi-channel-wise scale update and timestep-aware scale update to optimize model performance for downstream tasks.

\section{Experiments}
In this section, we evaluate the performance of the TuneQDM across various tasks (\ie, single-, multi-subject and unconditional generation. Unless specified otherwise, we utilize the DDIM sampler with $\eta=0$ and 50 steps for single-subject generation, and 100 steps for multi-subject and unconditional generation. Implementation details can be found in the supplementary material.

\subsection{Main results}

\begin{table}[t]
  \caption{\textbf{Quantitative comparison of single-subject generation.}}
  \label{tab:quantitative results-single}
\centering \resizebox{\textwidth}{!}
{
  \begin{tabular}{lccccccc}
    \toprule
    Method & Bits(W) & Size & \# Params & DINO-I & CLIP-I & CLIP-T \\
    \midrule
    Full prec.      & 32    & 3.20GB              & 859M              & 0.431 & 0.746     & 0.316 \\
    \midrule
    Baseline        & 4     & 0.40GB {\scriptsize + 1.32MB}     & 0.33M          & 0.519   & 0.787     & 0.313 \\
    TuneQDM         & 4     & 0.40GB {\scriptsize + 2.48MB}     & 0.62M          & 0.551 \color{blue}{\scriptsize $(+6.16\%)$}   & 0.802 \color{blue}{\scriptsize $(+1.91\%)$}    & 0.306 \color{red}{\scriptsize $(-2.23\%)$}\\
    \midrule
    Baseline        & 8     & 0.80GB {\scriptsize + 1.32MB}     & 0.33M          & 0.581   & 0.824     & 0.300 \\
    TuneQDM         & 8     & 0.80GB {\scriptsize + 2.48MB}     & 0.62M          & 0.578 \color{red}{\scriptsize $(-0.52\%)$}  & 0.816  \color{red}{\scriptsize $(-0.97\%)$}   & 0.307 \color{blue}{\scriptsize $(+2.33\%)$}\\
    \bottomrule
  \end{tabular}
}
\end{table}

\begin{figure}[t]
  \centering
  \includegraphics[width=\linewidth]{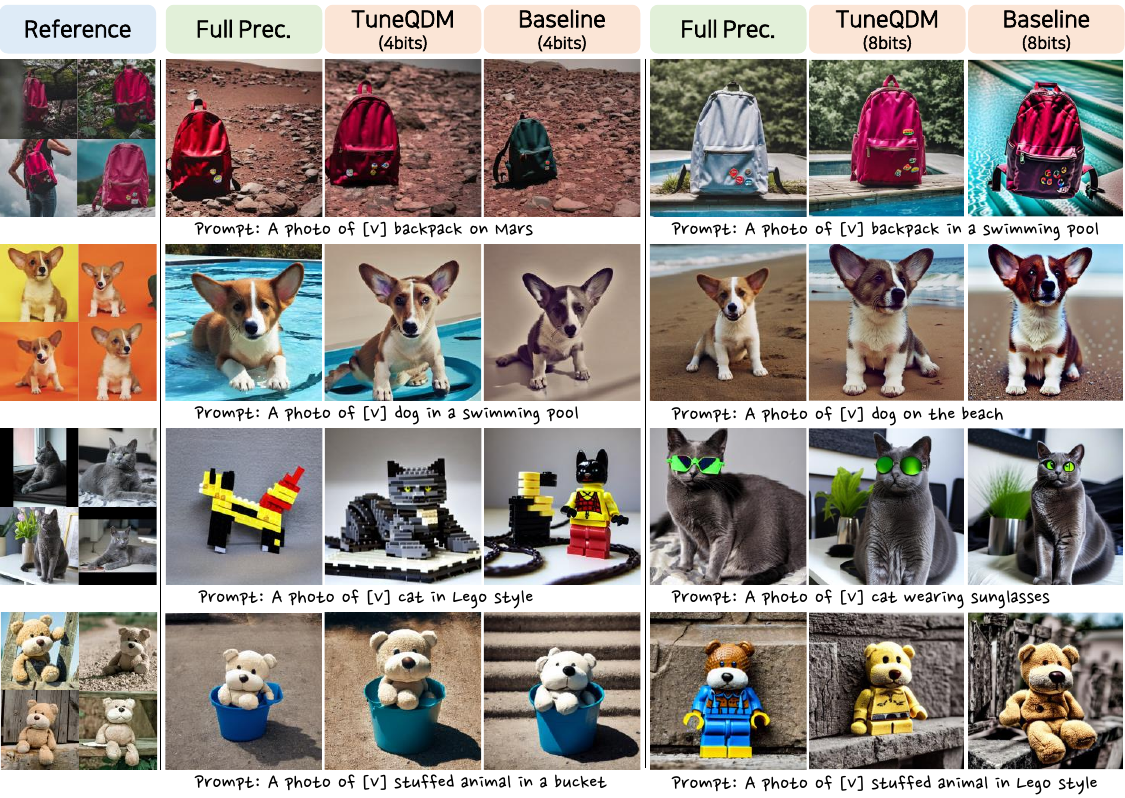}
  \caption{\textbf{Qualitative comparisons of single-subject generation.} We compared the \texttt{fp} model, TuneQDM, and baseline that fine-tuned on the target images. Subject fidelity and prompt fidelity were assessed for images generated by both TuneQDM and the baseline.}
  \label{fig:exp-single-subject}
\end{figure}

\noindent\textbf{Comparison on single-subject generation}.
Table \ref{tab:quantitative results-single} and Fig. \ref{fig:exp-single-subject} show the quantitative and qualitative comparison between our TuneQDM, \texttt{fp} and the \texttt{baseline} models. For quantitative evaluation, we assess both image and prompt fidelity. Image fidelity is measured using the CLIP \cite{hessel-etal-2021-clipscore}, and DINO \cite{oquab2023dinov2} image similarity. Prompt fidelity is evaluated using the CLIP text-to-image similarity. 

In the 8-bit setting, the difference between TuneQDM and the \texttt{baseline} model is minimal. However, in the 4-bit setting, TuneQDM outperforms the \texttt{baseline} model in subject fidelity while maintaining a similar level of prompt fidelity. Specifically, we highlight that, as mentioned in previous studies \cite{ruiz2023dreambooth, hao2023vico}, DINO-I better reflects prompt fidelity. This can be more clearly observed through qualitative comparison. 

As shown in Fig. \ref{fig:exp-single-subject}, TuneQDM consistently outperforms the \texttt{baseline} by generating images that more accurately reflect both subject features and prompts. In the 4-bit setting, TuneQDM significantly performs better in accurately representing subject features compared to the \texttt{baseline} (row 1, 3, 4). In row 2, TuneQDM effectively reflects the prompt (\ie, the swimming pool) into the image, unlike the \texttt{baseline} model.
Moreover, TuneQDM often produces higher quality images than \texttt{baseline}. Compared to the \texttt{fp} model, TuneQDM achieves similar results despite using a $\times 8$ compressed quantization model with much fewer parameters. % Additionally, in some cases the TuneQDM model outperforms the \texttt{fp} model.

\begin{table}
  \caption{\textbf{Quantitative comparison of multi-subject generation. }}
  \label{tab:quantitative results-multi}
\centering \resizebox{\textwidth}{!}
{
  \begin{tabular}{lccccccc}
    \toprule
    Method & Bits(W) & Size & \# Params & DINO-I & CLIP-I & CLIP-T \\
    \midrule
    Full prec.      & 32    & 3.20GB              & 859M              & 0.345 & 0.706     & 0.304 \\
    \midrule
    Baseline        & 4     & 0.40GB {\scriptsize + 1.32MB}     & 0.33M          & 0.275   & 0.677     & 0.314 \\
    TuneQDM         & 4     & 0.40GB {\scriptsize + 2.48MB}     & 0.62M          & 0.276 \color{blue}{\scriptsize $(+0.36\%)$}    & 0.675 \color{red}{\scriptsize $(-0.30\%)$}    & 0.317 \color{blue}{\scriptsize $(+0.96\%)$} \\
    \midrule
    Baseline        & 8     & 0.80GB {\scriptsize + 1.32MB}     & 0.33M          & 0.330  & 0.704   & 0.286  \\
    TuneQDM         & 8     & 0.80GB {\scriptsize + 2.48MB}     & 0.62M          & 0.329  \color{red}{\scriptsize $(-0.30\%)$}  & 0.708  \color{blue}{\scriptsize $(+0.57\%)$}    & 0.295 \color{blue}{\scriptsize $(+3.15\%)$} \\
    \bottomrule
  \end{tabular}
}
\end{table}

\begin{figure}[h!]
  \centering
  \includegraphics[width=\linewidth]{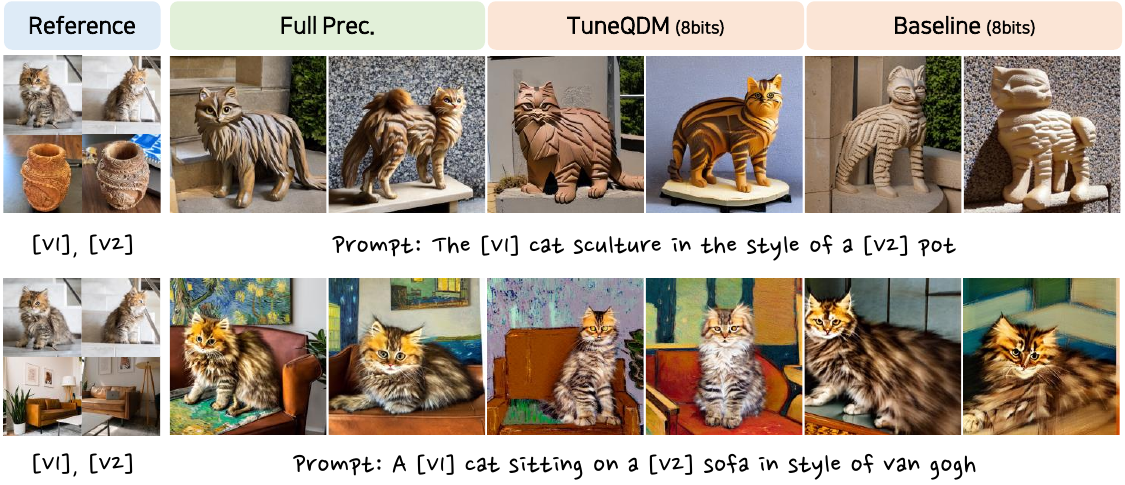}
  \caption{\textbf{Qualitative comparisons of multi-subject generation.} The images generated by TuneQDM exhibit excellent quality by capturing details effectively and reflecting the prompts and subject features accurately.
  }
  \label{fig:exp-multi-subject}
\end{figure}

\noindent\textbf{Comparison on multi-subject generation}
Table \ref{tab:quantitative results-multi} and Fig. \ref{fig:exp-multi-subject} summarize the quantitative and qualitative comparison results. The \texttt{baseline} model frequently fails to reflect the features of the subjects and the prompt. In contrast, TuneQDM successfully captures both the subjects' features and the prompt. As shown in Table \ref{tab:quantitative results-multi}, while DINO-I and CLIP-I scores are nearly identical, TuneQDM shows a higher CLIP-T score. However, both the \texttt{baseline} and TuneQDM show a performance drop compared to the \texttt{fp} model. 

In Fig. \ref{fig:exp-multi-subject}, the differences between each model are more evident. In row 1, TuneQDM effectively captures the features of both the ``[V1] cat'' and the ``[V2] pot'', whereas the \texttt{baseline} fails to represent the characteristics of the ``[V1] cat''. In row 2, the \texttt{baseline} does not depict the ``[V2] sofa'' at all, while TuneQDM successfully represents the features of both the ``[V1] cat'' and the ``[V2] sofa''. However, there were some cases where TuneQDM cannot perfectly reflect both prompt and subject fidelity. In multi-subject generation, which requires learning two concepts simultaneously, there was a slight performance gap compared to the \texttt{fp} model, unlike in single-subject generation. More qualitative results can be found in the supplementary material.

\begin{table} % [h]
\caption{\textbf{Performance comparison of fine-tuned quantized diffusion models on CIFAR-10 32 $\times$ 32. }} 
\centering
    {
    \resizebox{0.65\columnwidth}{!}
    {%
        \begin{tabular}{lcccccc}
        \toprule
        \textbf{Model} & \textbf{Bits} & \textbf{Model Size} & \textbf{\# Params} & \textbf{IS} & \textbf{FID} \\
        \midrule
        Full Prec.          & 32 & 143.1MB  & 35.8M   & 9.00 & 4.53 \\
        \midrule
        Q-Diffusion    & 8  & 35.8MB   & -       & 8.97 & 4.45 \\
        + QLoRA (r=32)   & 8  & 35.8MB   & 8.64M  & 9.03 & 4.30 \\
        + QLoRA (r=2)    & 8  & 35.8MB   & 0.57M  & 9.03 & 4.15 \\
        + Baseline       & 8  & 35.8MB   & 0.03M  & 8.96 & 4.39 \\
        + TuneQDM        & 8  & 35.8MB   & 0.13M  & \textbf{9.17} & \textbf{3.80} \\
        \bottomrule
        \end{tabular}
    }
    \label{tab:model_comparison}
    }
\end{table}

\noindent\textbf{Comparison on unconditional generation}
To evaluate the performance of TuneQDM for tasks other than personalization, we conducted fine-tuning to enhance the original purpose of the quantized diffusion model, similar to the previous study \cite{he2023efficientdm}. For this, we tested the performance on unconditional generation using the CIFAR-10 dataset. For this task, we compared our method with the baseline and QLoRA. The performance of diffusion models is evaluated with Inception Score(IS) and Fréchet inception distance(FID). 
As shown in Table \ref{tab:model_comparison}, the baseline performed similarly to QLoRA, while TuneQDM outperformed QLoRA in both metrics with fewer parameters.

\begin{figure}[t]
  \centering
  \begin{subfigure}{0.48\linewidth}
    \includegraphics[width=\linewidth]{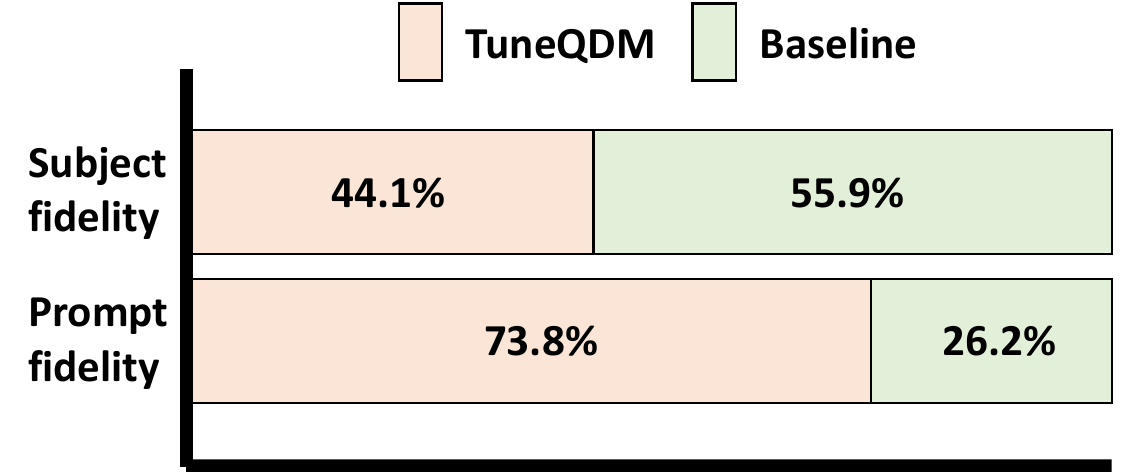}
    \caption{Single-subject generation}
    \label{fig:user-a}
  \end{subfigure}
  \hfill
  \begin{subfigure}{0.48\linewidth}
    \includegraphics[width=\linewidth]{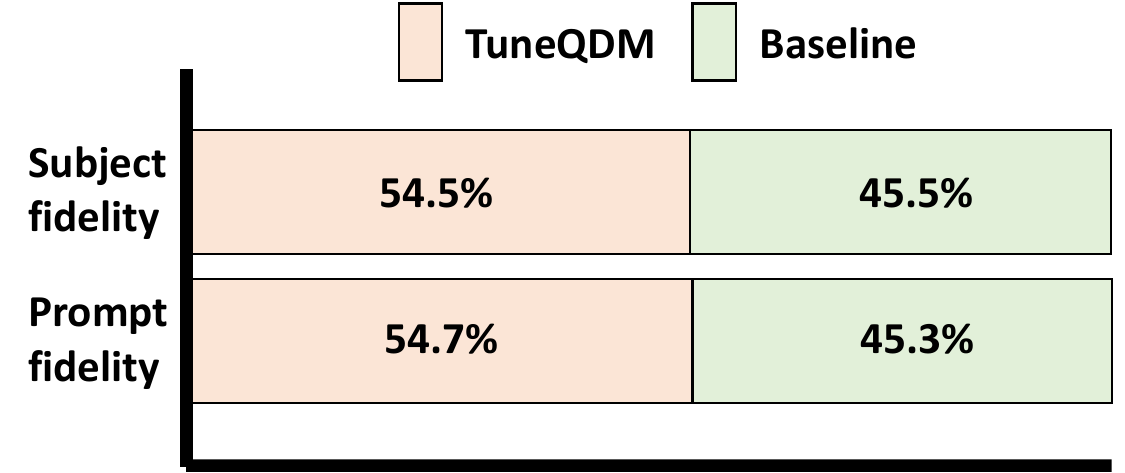}
    \caption{Multi-subject generation}
    \label{fig:user-b}
  \end{subfigure}
  \caption{\textbf{User study.} All models are 4-bit quantized diffusion models.}
  \label{fig:user-study-main}
\end{figure}

\noindent\textbf{User study}. \label{user_study_subsubsec}
To evaluate prompt and subject fidelity accurately, we conducted a user study. As shown in Fig. \ref{fig:user-study-main}, TuneQDM exhibits slightly lower subject fidelity but significantly higher prompt fidelity compared to the baseline in single-subject generation. In a multi-subject generation, TuneQDM shows slightly higher fidelity in both subject and prompt fidelity.
As known from previous research \cite{kumari2023customdiffusion}, prompt and subject fidelity generally have a trade-off relationship. Therefore, achieving better performance in both prompt and subject, even with slight differences, signifies a meaningful performance improvement. Experimental details can be found in the supplementary material.

\begin{table}[t]
\caption{\textbf{Ablation study.} Results showing the impact of MCSU and TAS on IS and FID.}
\centering
    {
    \resizebox{0.5\columnwidth}{!}
    {
        \begin{tabular}{lccccc}
        \toprule
        \textbf{Model} & \textbf{Bits} & \textbf{MCSU} & \textbf{TAS} & \textbf{IS} & \textbf{FID} \\
        \midrule
        Full Prec. & 32 & -     & - & 9.00 & 4.53 \\
        \midrule
        Baseline      & 8  & \xmark & 1 & 8.96 & 4.39    \\
        \midrule
                       & 8 & \xmark & 2 & \textbf{9.19} & 4.24    \\
                       & 8 & \xmark & 4 & 9.03 & 4.25    \\
        TuneQDM        & 8 & \checkmark  & 1 & 8.97 & 4.33    \\
                       & 8 & \checkmark  & 2 &\underline{9.17} & \textbf{3.80}    \\
                       & 8 & \checkmark  & 4 & 9.02 & \underline{4.15}    \\
        \bottomrule
        \end{tabular}
    }
    \label{tab:ablation_study}
    }
\end{table}

\subsection{Analysis}

\noindent\textbf{Ablation study}.
To assess the impact of each component, we conducted ablation studies. As shown in Table \ref{tab:ablation_study}, applying multi-channel-wise scale update (MCSU) and Timestep-aware scale update (TAS) with two experts to the baseline achieved the best performance in terms of FID. There was an improvement in performance when each component was applied.

\begin{figure}[t!]
  \centering
  \includegraphics[width=\linewidth]{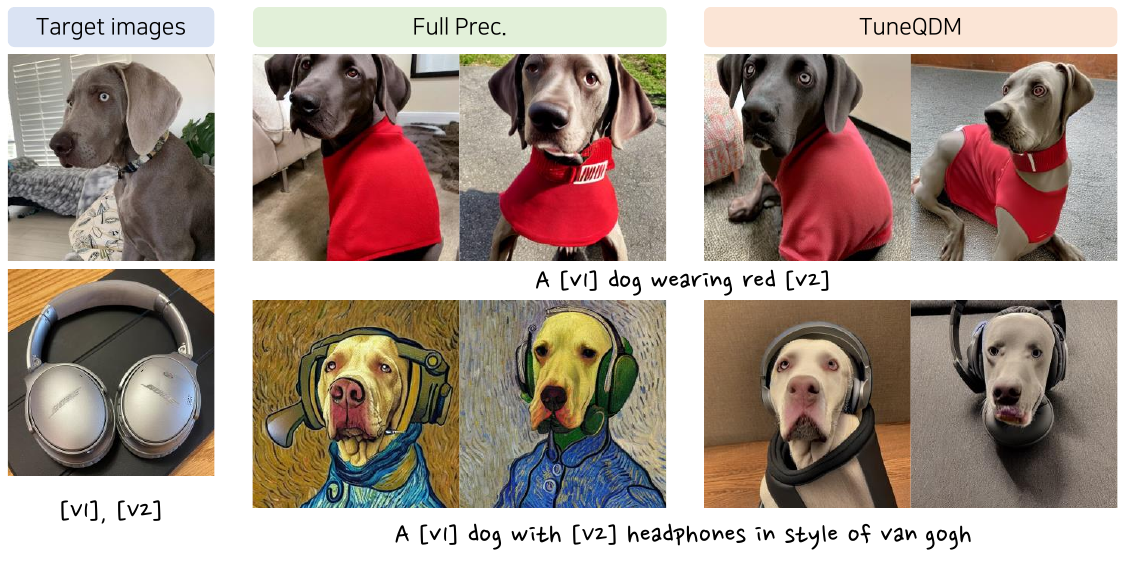}
  \caption{\textbf{Failure cases}. Above: Both the \texttt{fp} and TuneQDM models fail to accurately reflect the prompt. Below: The \texttt{fp} model successfully captures both the subject and prompt, while TuneQDM does not.
  }
  \label{fig:limitation}
\end{figure}

\noindent\textbf{Limitation}.
As shown in Fig \ref{fig:limitation}, our method has several failure cases. In a multi-subject generation, there were some cases where only one subject was reflected, or the prompt was not reflected. The first row shows cases where the full precision model also failed to reflect the subjects (case 1). As shown in the second row, there were cases where subjects were well reflected in the full precision model but not in the quantized model (case 2). Case 1 can be attributed to the limitations of the personalization methodology, whereas case 2 specifically occurs when using the quantized model, indicating the need for additional solutions to address this problem.

\section{Conclusion}

This paper addressed the problem of the fine-tuning of quantized diffusion models for the first time. Inspired by the unique characteristics of diffusion models, we proposed TuneQDM, a memory-efficient fine-tuning method for quantized models. Our approach represents scales as separable functions to account for weight update patterns and customizes quantization scales for distinct intervals, effectively enhancing model capacity with minimal memory overhead. TuneQDM achieves both parameter efficiency and a substantial reduction in memory footprint during fine-tuning. Our experimental results demonstrate that our method achieves high subject fidelity and prompt fidelity while mitigating overfitting, significantly outperforming the baseline approach.

We believe fine-tuning the low-precision vision foundation models, the quantized diffusion models, holds great potential for diverse computer vision applications, alleviating slow inference and resource demands. This work can facilitate the practical deployment of diffusion models in real-world computer vision scenarios.

\section*{Acknowledgements}
This research was supported by the Basic Science Research Program through the National Research Foundation of Korea (NRF) funded by the MSIP (NRF-2022R1A2C3011154, RS-2023-00219019, RS-2023-00240135), KEIT grant funded by the Korea government (MOTIE) (No. 2022-0-00680, 2022-0-01045, 2021-0-02068, Artificial Intelligence Innovation Hub), the IITP grant funded by the Korea government (MSIT) (RS-2019-II190075, Artificial Intelligence Graduate School Program(KAIST)).

%% file: 2.supp.tex
\clearpage
\setcounter{page}{1}

\appendix
\renewcommand{\thetable}{A.\arabic{table}}
\renewcommand{\thefigure}{A.\arabic{figure}}
\renewcommand{\thealgorithm}{A.\arabic{algorithm}}

\setcounter{section}{0}
\setcounter{table}{0}
\setcounter{figure}{0}
\setcounter{algorithm}{0}

% ---------------------------------------------------------------
\title{Memory-Efficient Fine-Tuning \\ for Quantized Diffusion Model \\ (Supplementary Material)} 

\titlerunning{TuneQDM}

\author{Hyogon Ryu \and
Seohyun Lim \and
Hyunjung Shim}

\authorrunning{Ryu et al.}

\institute{Korea Advanced Institute of Science and Technology (KAIST)\\
\email{\{hyogon.ryu, seohyunlim, kateshim\}@kaist.ac.kr}}

\maketitle

\section{Implementation detail}
All experiments were reproduced by our implementation based on \texttt{Diffusers}\footnote{\url{https://github.com/huggingface/diffusers}} library. For quantized checkpoint, we use q-diffusion\footnote{\url{https://github.com/Xiuyu-Li/q-diffusion}}'s official checkpoint. Experiments regarding to Full-precision Dreambooth\footnote{\url{https://huggingface.co/docs/diffusers/training/dreambooth}} and Custom Diffusion\footnote{\url{https://huggingface.co/docs/diffusers/training/custom_diffusion}} are conducted based on \texttt{Diffusers} official implementation without any editing.
Pseudo code for multi channel-wise scale update is available in Algorithm \ref{supp:alg:linear}

\subsection{Hyperparameter}
For single-subject generation inference, we utilize a guidance scale of 7.5 and set eta to 0, with a DDIM step of 50. For multi-subject generation, we adjust the parameters to a guidance scale of 5.0, eta of 1.0, and a DDIM step of 100. This configuration is for preserving the default setting.

In the case of multi-subject generation, prior loss is employed. However, for single-subject generation, prior loss is excluded, as the quantized model often fails to fine-tune for the target subject in almost cases.

\begin{table}
\caption{\textbf{Learning rate.} The values of full precision are same as the default setting mentioned in the original paper, as discussed in the main paper.
  }
  \label{tab:learning rate}
\centering
{
  \begin{tabular}{lcccc}
    \toprule
    Method &  \quad Full prec. \quad   &  \quad 4bits \quad   &  \quad 8bits \quad   \\
    \midrule
    Dreambooth      &   \quad 5e-6   &  \quad 3e-5             &  \quad 3e-6  \\
    CustomDiffusion        &  \quad 1e-5     &  \quad 1e-5     &  \quad 1e-5  \\
    \bottomrule
  \end{tabular}
}
\end{table}

We used a batch size of 1 for Dreambooth and 2 for Custom Diffusion. We generated the images with train iteration 400 and 800, then selected the better one. For fair comparison, except for the learning rate, all hyperparameters are set to the same values for both the full precision and quantized models. Learning rates are displayed in Table \ref{tab:learning rate}.
We searched for the best setting for the baseline and then applied it to TuneQDM as well. Since we didn't search for the best settings for TuneQDM, there might be a possibility of slight performance improvement through hyperparameter search.

\subsection{Metric}
To measure subject fidelity, we evaluated DINO-I\cite{oquab2023dinov2} and CLIP-I\cite{hessel-etal-2021-clipscore} scores, while for prompt fidelity, we measured CLIP-T scores. The CLIP encoder used ViT-B/32, and DINO-I utilized DINOv2 ViT-S/14.
DINO, being trained via self-supervised methods, is known to measure differences well compared to the CLIP image encoder when given the similar type of subject.

\subsection{Training loss} 
To fine-tune Stable Diffusion, we utilize the same loss function as employed in DreamBooth and Custom Diffusion. The loss is defined as the weighted sum of the prior-preservation loss and the simple diffusion loss. The loss function can be expressed as follows:

\begin{align}
    \mathcal{L} &= \mathbb{E}_{z, c, \epsilon, t}[||\hat \epsilon_\theta(z, c) - \epsilon ||^2] + \lambda \mathcal{L}_{\text{prior}}, \\
    \mathcal{L}_{\text{prior}} &= \mathbb{E}_{z_{\text{pr}}, c_{\text{pr}}, \epsilon, t}[||\hat \epsilon_\theta(z_{\text{pr}}, c_{\text{pr}}) - \epsilon ||^2].
\end{align}

Here, $\mathcal{L}$ represents the total loss, $\mathcal{L}_{\text{prior}}$ denotes the prior-preservation loss, $\hat \epsilon_\theta(z, c)$ and $\hat \epsilon_\theta(z_{\text{pr}}, c_{\text{pr}})$ are the generated noise vectors corresponding to the target images and prior examples, respectively. $z$ and $c$ represent the target image latents and text embeddings, $z_{\text{pr}}$ and $c_{\text{pr}}$ represent the latent and text embeddings for the prior examples, $\epsilon$ represents the ground truth noise vector, $\lambda$ is a weighting coefficient, and $t \sim \mathcal{N}(1, T)$ represents the diffusion timestep.

By optimizing the aforementioned loss function during fine-tuning, the adapted diffusion model becomes capable of generating single and multi-subject images tailored to specific user preferences or input text prompts.

\section{Additional results}
\subsection{Quantitative Results}

Table \ref{tab:quantitative results-appendix-single} and \ref{tab:quantitative results-appendix-multi} present the quantitative results for each task, evaluated using DINO-I, CLIP-I, and CLIP-T scores. While the differences in CLIP-T scores are negligible, significant differences exist between TuneQDM and the baseline in terms of DINO-I and CLIP-I scores. However, as mentioned in the main paper, measuring subject- and prompt fidelity using DINO and CLIP scores is inaccurate. Therefore, it is necessary to evaluate through qualitative results and user studies.

\begin{table}
  \caption{\textbf{Quantitative Comparison of single-subject generation.} 
TuneQDM$^{\star}$ initializes the multi-channel-wise scale from $\mathcal{N}(0, 0.01)$.}
  \label{tab:quantitative results-appendix-single}
\centering \resizebox{\textwidth}{!}
{
  \begin{tabular}{lccccccc}
    \toprule
    Method & Bits(W) & Size & \# Params & DINO-I & CLIP-I & CLIP-T \\
    \midrule
    Full prec.      & 32    & 3.20GB              & 859M              & 0.431 & 0.746     & 0.316 \\
    \midrule
    Baseline        & 4     & 0.40GB {\scriptsize + 1.32MB}     & 0.33M          & 0.519   & 0.787     & 0.313 \\
    TuneQDM         & 4     & 0.40GB {\scriptsize + 2.48MB}     & 0.62M          & 0.551 \color{blue}{\scriptsize $(+6.16\%)$}   & 0.802 \color{blue}{\scriptsize $(+1.91\%)$}    & 0.306 \color{red}{\scriptsize $(-2.23\%)$}\\
    \midrule
    Baseline        & 8     & 0.80GB {\scriptsize + 1.32MB}     & 0.33M          & 0.581   & 0.824     & 0.300 \\
    TuneQDM         & 8     & 0.80GB {\scriptsize + 2.48MB}     & 0.62M          & 0.584 \color{blue}{\scriptsize $(+0.52\%)$}  & 0.830 \color{blue}{\scriptsize $(+0.73\%)$}    & 0.298 \color{red}{\scriptsize $(-0.67\%)$}\\
    TuneQDM$^\star$         & 8     & 0.80GB {\scriptsize + 2.48MB}     & 0.62M          & 0.578 \color{red}{\scriptsize $(-0.52\%)$}  & 0.816  \color{red}{\scriptsize $(-0.97\%)$}   & 0.307 \color{blue}{\scriptsize $(+2.33\%)$}\\
    \bottomrule
  \end{tabular}
}
\end{table}

\begin{table}
  \caption{\textbf{Quantitative Comparison of multi-subject generation. }
TuneQDM$^{\star}$ initializes the multi-channel-wise scale from $\mathcal{N}(0, 0.01)$.}
  \label{tab:quantitative results-appendix-multi}
\centering \resizebox{\textwidth}{!}
{
  \begin{tabular}{lccccccc}
    \toprule
    Method & Bits(W) & Size & \# Params & DINO-I & CLIP-I & CLIP-T \\
    \midrule
    Full prec.      & 32    & 3.20GB              & 859M              & 0.345 & 0.706     & 0.304 \\
    \midrule
    Baseline        & 4     & 0.40GB {\scriptsize + 1.32MB}     & 0.33M          & 0.275   & 0.677     & 0.314 \\
    TuneQDM         & 4     & 0.40GB {\scriptsize + 2.48MB}     & 0.62M          & 0.276 \color{blue}{\scriptsize $(+0.36\%)$}    & 0.675 \color{red}{\scriptsize $(-0.30\%)$}    & 0.317 \color{blue}{\scriptsize $(+0.96\%)$} \\
    \midrule
    Baseline        & 8     & 0.80GB {\scriptsize + 1.32MB}     & 0.33M          & 0.330  & 0.704   & 0.286  \\
    TuneQDM         & 8     & 0.80GB {\scriptsize + 2.48MB}     & 0.62M          & 0.329  \color{red}{\scriptsize $(-0.30\%)$}  & 0.708  \color{blue}{\scriptsize $(+0.57\%)$}    & 0.295 \color{blue}{\scriptsize $(+3.15\%)$} \\
    TuneQDM$^\star$         & 8     & 0.80GB {\scriptsize + 2.48MB}     & 0.62M          & 0.329  \color{red}{\scriptsize $(-0.30\%)$}  & 0.705   \color{blue}{\scriptsize $(+0.14\%)$}   & 0.293 \color{blue}{\scriptsize $(+2.45\%)$} \\
    \bottomrule
  \end{tabular}
}
\end{table}

\subsection{Explanation about full precision's DINO-I, CLIP-I score}

\begin{figure}[h!]
  \centering
  \includegraphics[width=0.8\linewidth]{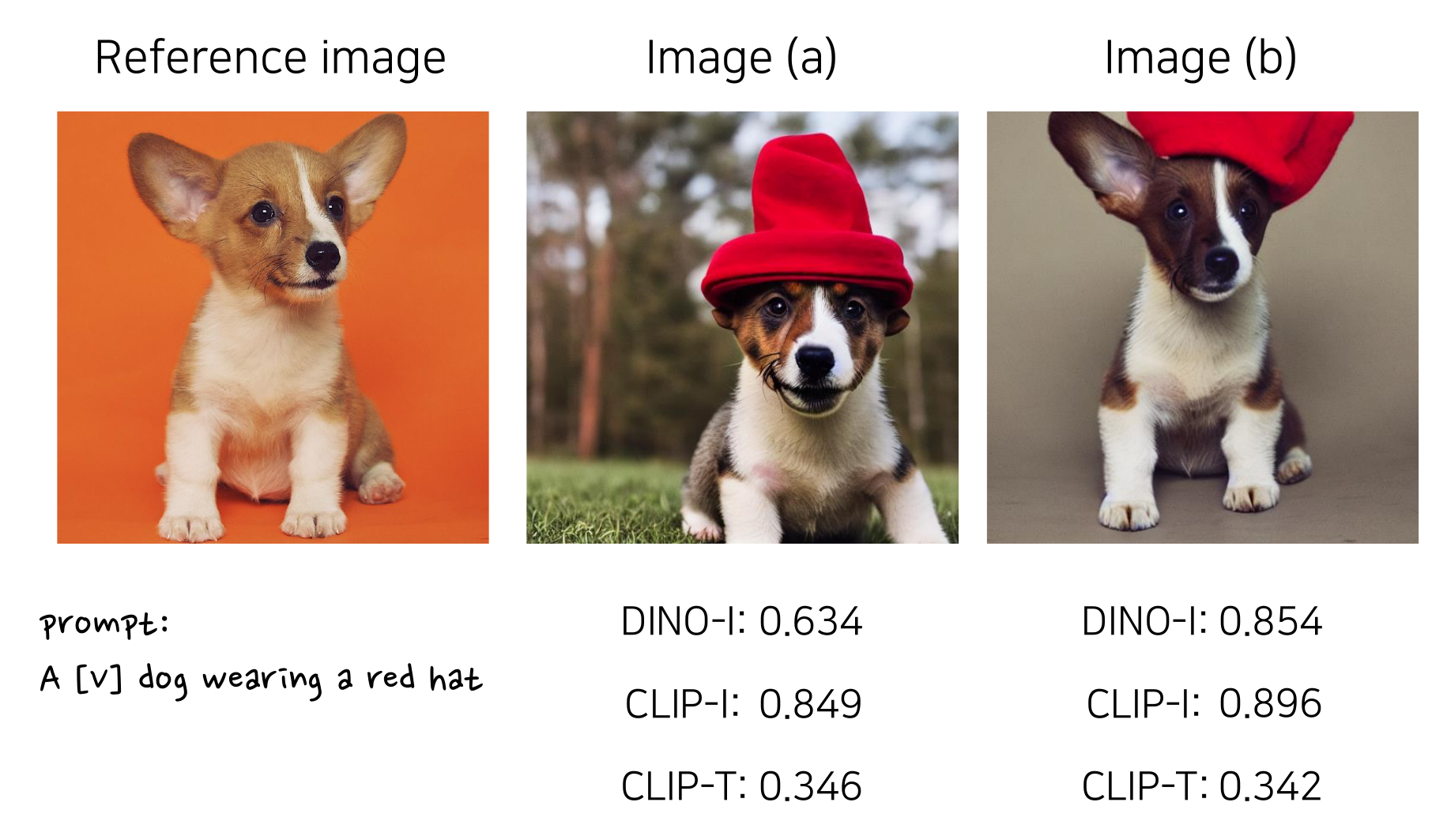}
  \caption{\textbf{Limitation of subject-fidelity metrics} 
  }
  \label{fig:metric-background}
\end{figure}

The DINO-I and CLIP-I scores are easily influenced by some components unrelated to subject fidelity. In Fig \ref{fig:metric-background}., despite both the (a) and (b) images effectively reflecting the features of the subject, there are significant differences in the DINO-I and CLIP-I scores. This difference occurred because the similarity of the background and subject's pose to the reference image had an effect on the score.
In the case of the full precision model, various components unrelated to the prompt (\eg background or subject pose) exhibited diversity, resulting in lower scores compared to the quantized model, as illustrated in the table.
Thus, evaluating whether the subject's features are well-reflected through CLIP-I and DINO-T scores is hard. Therefore, as repeatedly mentioned, it is essential to focus on qualitative results or conduct a user study to evaluate the performance accurately.

\subsection{Inference speed}
Our method focuses on memory efficiency through weight-only quantization. When examining its impact on inference speed, two aspects must be considered. First, quantizing weights to 4 bits reduces the cost of memory allocation on the GPU to $\frac{1}{4}$. However, the overhead of the dequantization process will slow down the operations such as matrix multiplication. Therefore, to increase inference speed through weight-only quantization, it is essential to verify if the actual speed improvement occurs by balancing memory and computational efficiency.
As noted by other studies \cite{lin2024awq}, considering the increasing size of recent models and the batch sizes in practical use scenarios are often 1 or 2, weight-only quantization can indeed be expected to improve inference speed.

Since implementing custom kernels for all layers of Stable Diffusion is challenging, we created a simple benchmark to test the inference time specifically for the linear layers where TuneQDM was applied. For multiplication operations, we used the GEMM kernel and conducted experiments on an A6000 GPU. As shown in Table \ref{tab:infer-speed}, both the baseline and TuneQDM were faster compared to full-precision and half-precision settings. However, the additional multiplication operations made TuneQDM slightly slower than the baseline.

\begin{table}
  \caption{\textbf{Inference speed comparison.}}
  \label{tab:infer-speed}
\centering 
{
  \begin{tabular}{lccc}
    \toprule
    Method & Bits(W) & Time \\
    \midrule
    full prec. & 32 & 15.60 s \\
    half prec. & 16 & 8.88 s \\
    \midrule
    Baseline & 4 & 6.94 s \\
    TuneQDM & 4 & 6.99 s \\
    \bottomrule
  \end{tabular}
}
\end{table}

\subsection{Additional qualitative results}

\begin{figure}
  \centering
  \includegraphics[width=\linewidth]{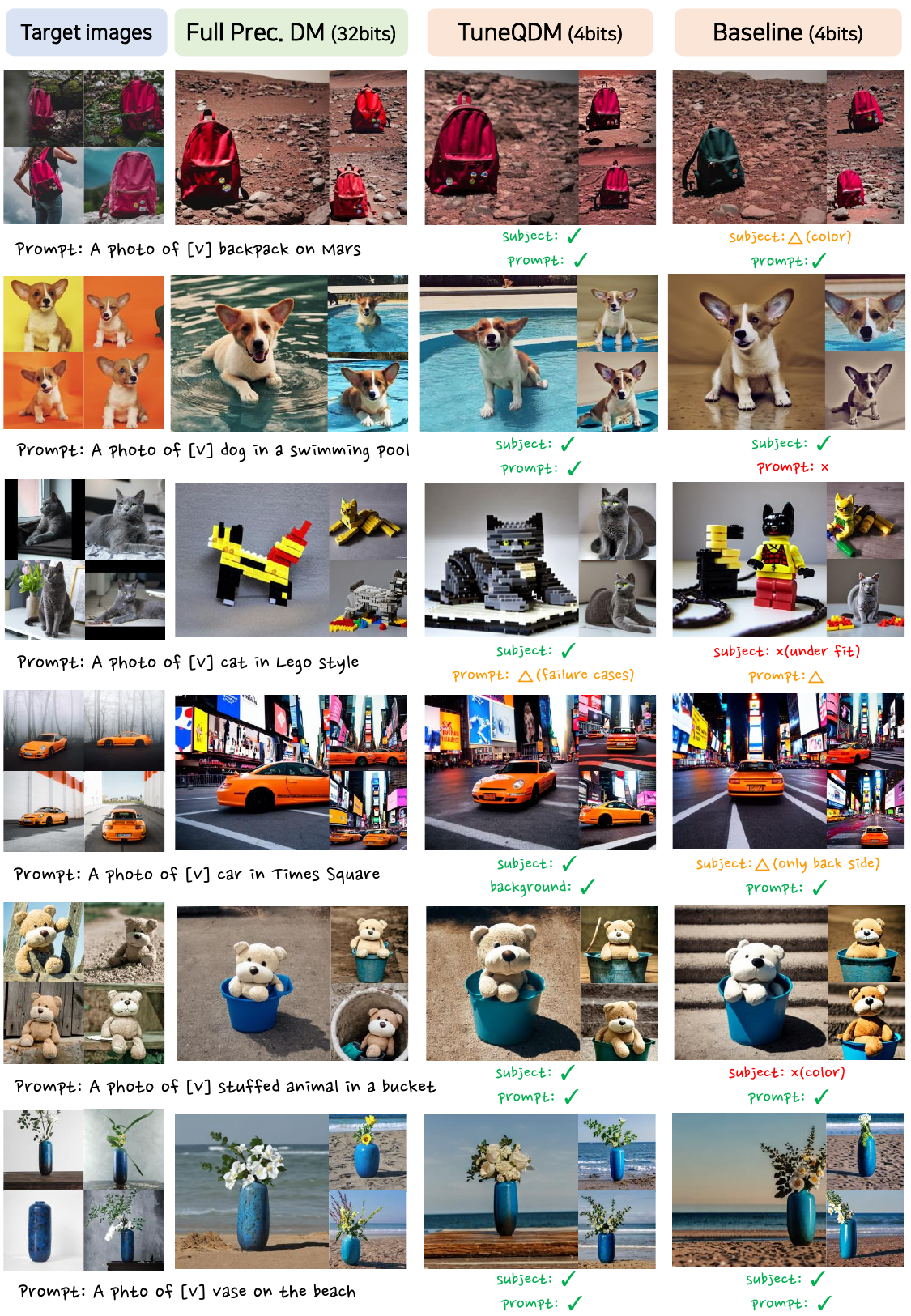}
  \caption{\textbf{Qualitative results of single-subject generation, 4bits} 
  }
  \label{fig:qual-single-4bits}
\end{figure}

\begin{figure}
  \centering
  \includegraphics[width=\linewidth]{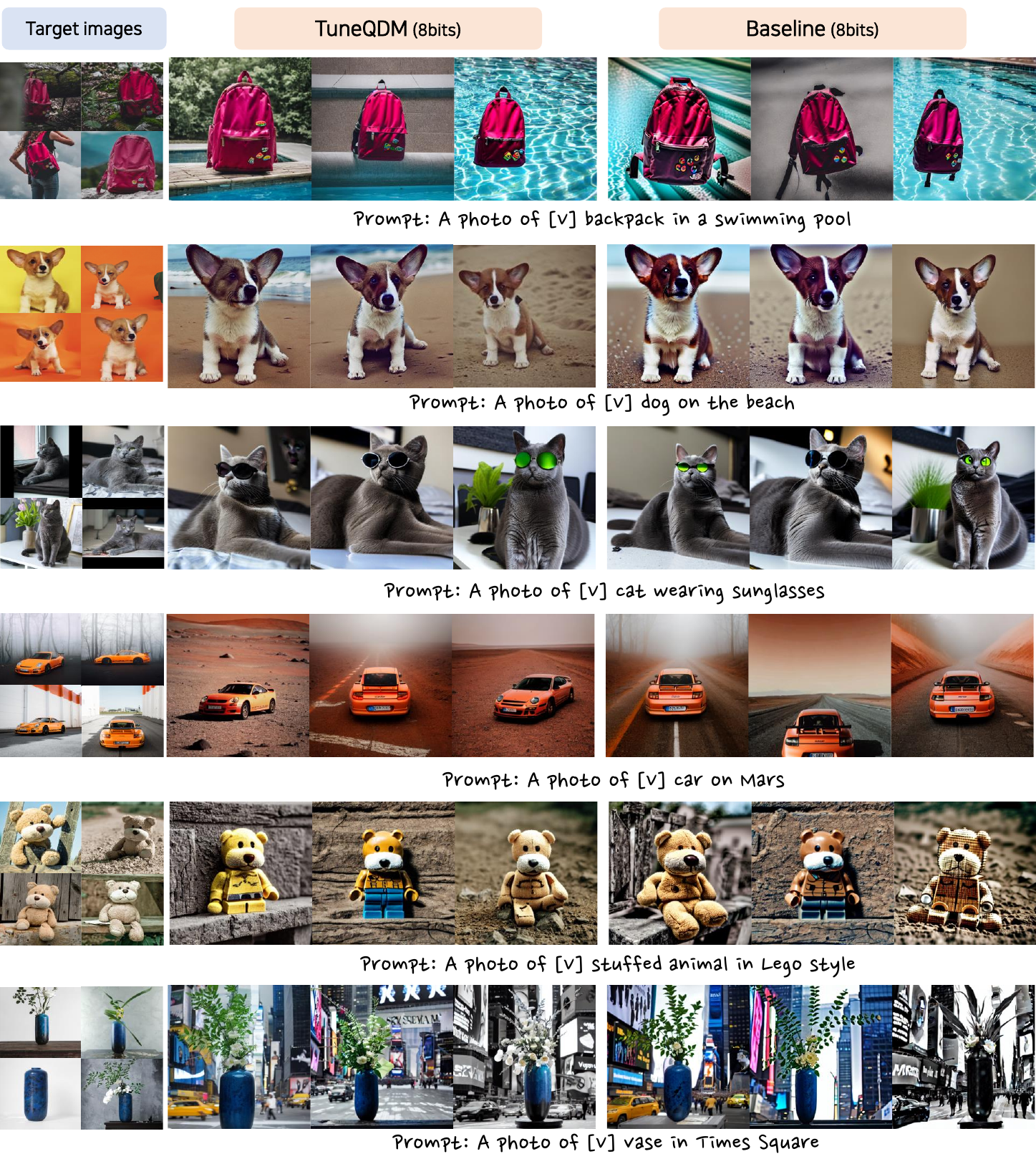}
  \caption{\textbf{Qualitative results of single-subject generation, 8bits} 
  }
  \label{fig:qual-single-8bits}
\end{figure}

\begin{figure}
  \centering
  \includegraphics[width=\linewidth]{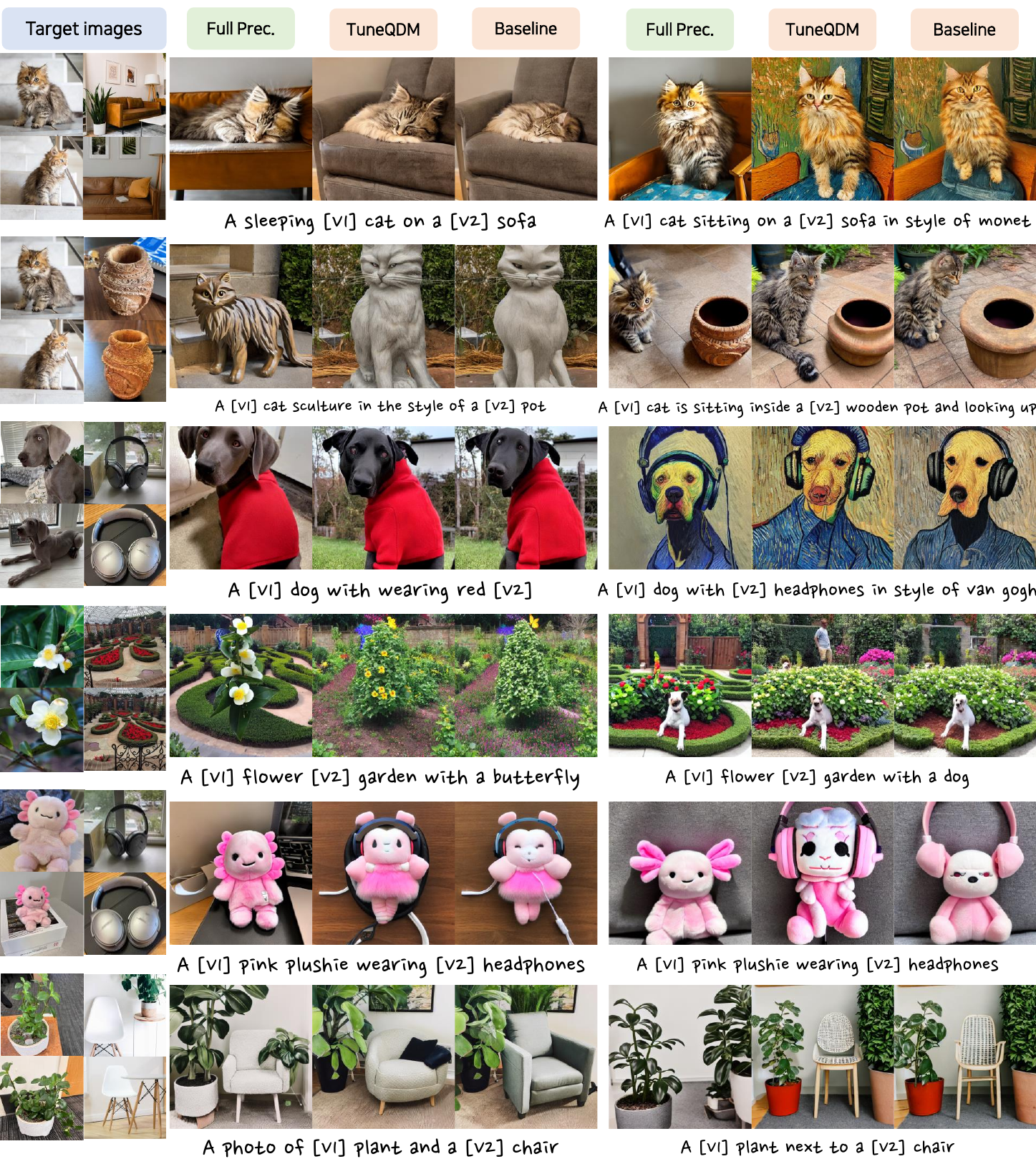}
  \caption{\textbf{Qualitative results of multi-subject generation, 4bits} 
  }
  \label{fig:qual-multi-4bits}
\end{figure}

\begin{figure}
  \centering
  \includegraphics[width=\linewidth]{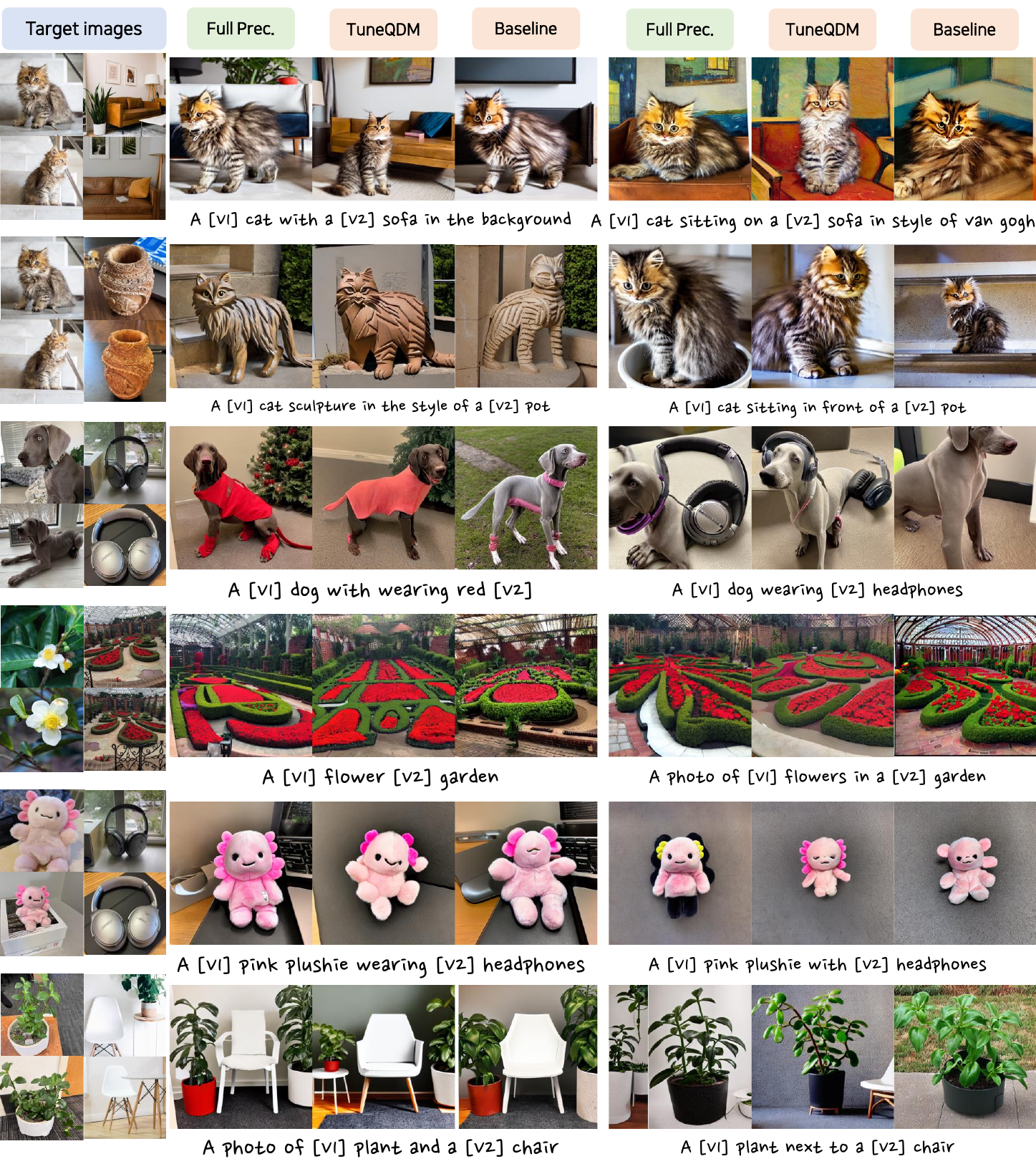}
  \caption{\textbf{Qualitative results of single-subject generation, 8bits} 
  }
  \label{fig:qual-multi-8bits}
\end{figure}

Fig. \ref{fig:qual-single-4bits} \ref{fig:qual-single-8bits}, \ref{fig:qual-multi-4bits}, and \ref{fig:qual-multi-8bits} respectively represent the qualitative results of single-subject generation with an 8-bit quantized model, multi-subject generation with 4-bit quantized model, and multi-subject generation with an 8-bit quantized model.

In Fig. \ref{fig:qual-single-8bits}, it can be seen that TuneQDM produces images that reflect both the subject and prompt better than the baseline. In particular, in rows 1, 4, and 5, the prompt is reflected much more harmoniously than in the baseline. While generating images that reflect the content of the prompt, as seen in the rightmost example in row 1, unnatural images can also be generated, but TuneQDM generates such unnatural images less frequently. In the case of row 6, both TuneQDM and the baseline did not produce satisfactory results.

For multi-subject generation, the overall quality of the generated images is unsatisfactory. This was influenced by the poor performance of the Full Precision model. Except for the cases where the cat was used (rows 1 and 2 in Fig. \ref{fig:qual-multi-4bits} and \ref{fig:qual-multi-8bits}), our experiments did not produce satisfactory results even when the full precision model was used for fine-tuning. We conducted experiments with the original codebase without any modifications when fine-tuning the full precision model.

Fig. \ref{fig:qual-multi-4bits} shows the results of multi-subject generation using a 4-bit quantized model. TuneQDM tends to be intermediate between Full Precision and the baseline. However, significant differences occur in cases where the presence or absence of subjects changes between the full precision model and the quantized model, as shown in rows 4 and 5.

Fig. \ref{fig:qual-multi-8bits} shows the results of multi-subject generation using an 8-bit quantized model. Similar to the 4-bit results, the 8-bit results show a similar trend. In particular, in rows 1 and 2, TuneQDM shows better performance than the baseline, and in the remaining rows, TuneQDM produces images closer to Full Precision than the baseline.

\section{User study details}

\begin{figure}
  \centering
  \includegraphics[width=\linewidth]{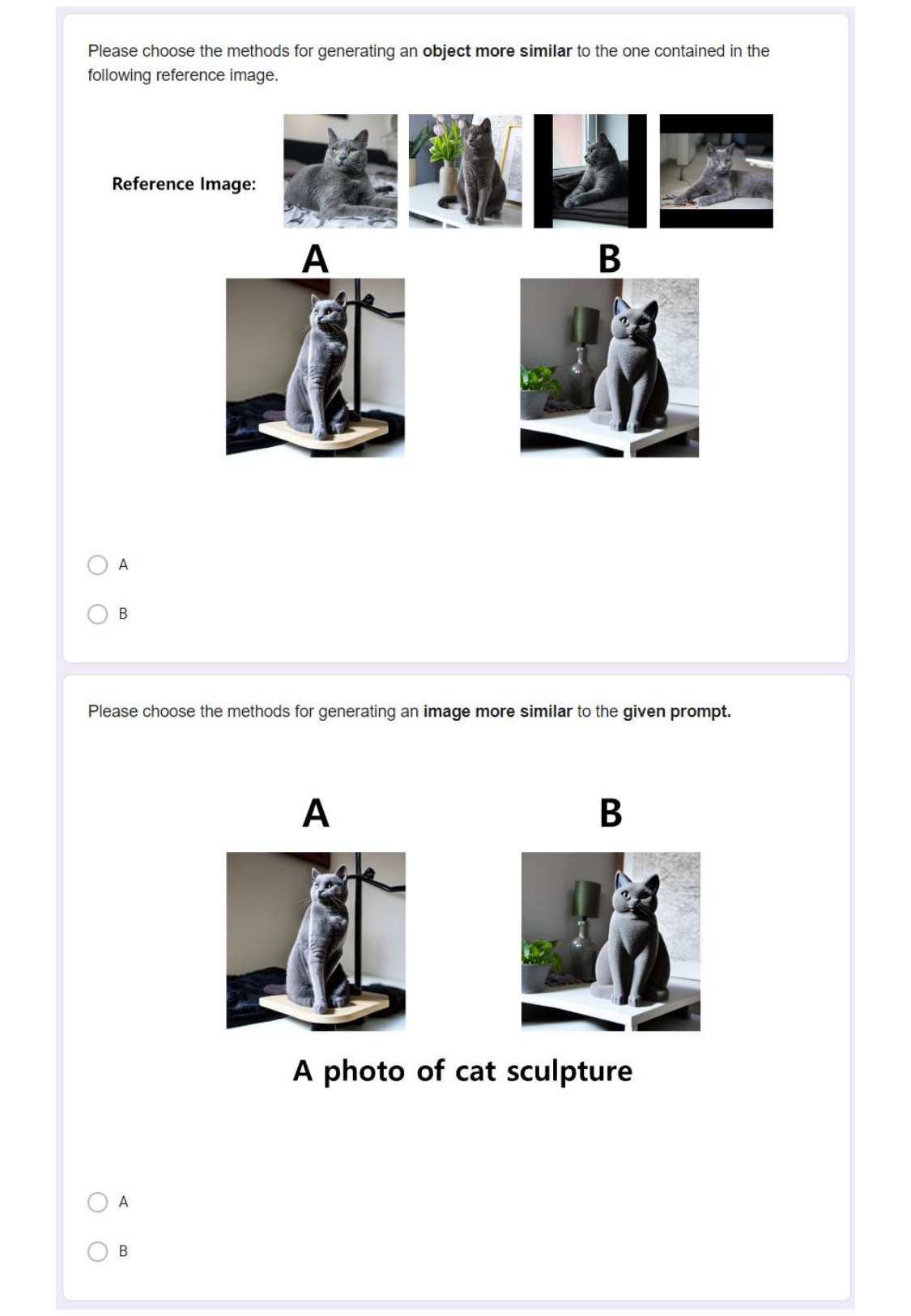}
  \caption{\textbf{example of the survey} 
  }
  \label{fig:user-study-example}
\end{figure}

We conducted a survey with a total of 86 questions to 45 participants. The survey focused on subject fidelity and prompt fidelity, comparing the baseline and TuneQDM to determine the preferred method. 56 questions were about single-subject generation, and 30 questions were about multi-subject generation. Baseline and TuneQDM were compared using the same configuration. An example of the survey is shown in Fig \ref{fig:user-study-example}.

\section{Discussions}

\subsection{Low-bits settings}

Our approach was generally more effective at 4 bits than at 8 bits. As the low-bit setting decreased, the capacity of the quantized model decreased, and the performance improvement achievable with our approach was greater. This is because our goal is ultimately to increase the training capacity of the model by providing denoising roles and applying multi-channel-wise scale update methods.

\subsection{Limitations}

It has been observed that the performance of multi-subject generation is significantly lower compared to single-subject generation. This appears to be due to inherent limitations in stable diffusion. Previous research\cite{feng2022training} has shown that stable diffusion does not effectively process images of multiple concepts. As a result, the limitations observed in multi-subject image generation persisted even in quantized models, and overcoming them is difficult even with our approach.

\subsection{future work}

The application of prior preservation loss did not yield satisfactory results. It appeared that the capacity of the quantized model was insufficient to learn new concepts while preserving the prior. There is a need to explore methods that facilitate effective tuning while maintaining the prior.

After fine-tuning the quantized model, even with the same seed, the resulting images differed from those of the full precision model. Considering other parameter-efficient fine-tuning methods that produce similar images to full fine-tuning even after fine-tuning completion using the same seed, our approach seems to fine-tune in a somewhat different manner compared to fine-tuning the full precision model. Research into methods to fine-tune such that the results of fine-tuning the full precision model and the quantized model are similar is warranted.

\clearpage

\begin{algorithm*}
\caption{Pseudo-Code for multi channel-wise scale update applied on Linear layer, PyTorch-like}
\label{supp:alg:linear}
\definecolor{codeblue}{rgb}{0.25,0.5,0.5}
\definecolor{codekw}{rgb}{0.85, 0.18, 0.50}
\lstset{
 backgroundcolor=\color{white},
 basicstyle=\fontsize{7.5pt}{7.5pt}\ttfamily\selectfont,
 columns=fullflexible,
 breaklines=true,
 captionpos=b,
 commentstyle=\fontsize{7.5pt}{7.5pt}\color{codeblue},
 keywordstyle=\fontsize{7.5pt}{7.5pt}\color{codekw},
}
\begin{lstlisting}[language=python]

class TQLinear(nn.Module):
    def __init__(self, QuantParam: Dict[str, Dict[str, torch.tensor]], weight: torch.tensor, bias):
        '''
        :param in_features: size of each input sample
        :param out_features: size of each output sample
        :param weight: weight tensor (quantized : dtype should be int)
        :param bias: bias tensor
        :param kwargs: other parameters

        :param QuantParam: load from quantized checkpoint
        '''
        super(TQLinear, self).__init__()      

        self.weight = weight
        if bias != None:
            self.bias = bias
        else:
            self.register_parameter('bias', None)

        self.delta = QuantParam['delta']
        self.zero_point = QuantParam['zero_point']
        self.n_bits = QuantParam['n_bits']
        self.sym = QuantParam['sym']

        self.delta = nn.Parameter(self.delta)
        self.double_delta = nn.Parameter(torch.ones((1, self.weight.shape[1])))
        torch.nn.init.normal_(self.double_delta, mean=1.0, std=0.1)
    
    def forward(self, input, *args, **kwargs):        
        return F.linear(input, (self.weight-self.zero_point) * self.delta * self.double_delta , self.bias)
\end{lstlisting}
\end{algorithm*}

\clearpage